\pdfoutput=1

\documentclass[11pt,table,xcdraw,xspace]{article}

\usepackage{emnlp2022}

\usepackage{times}
\usepackage{latexsym}
\usepackage{amssymb}

\usepackage[T1]{fontenc}

\usepackage[utf8]{inputenc}
\usepackage{CJKutf8}

\usepackage{graphicx}
\usepackage{multirow}
\usepackage{multicol}

\usepackage[normalem]{ulem}
\useunder{\uline}{\ul}{}

\usepackage{xspace}

\usepackage{dblfloatfix}

\usepackage{microtype}

\usepackage[normalem]{ulem}

\usepackage[]{todonotes}

\title{Machine Translation Robustness to Natural Asemantic Variation}

\author{Jacob Bremerman \, Xiang Ren \, Jonathan May\thanks{\xspace\xspace Work done prior to JM joining Amazon.}\\
University of Southern California \\
Information Sciences Institute \\
\texttt{\{jbrem,xiangren,jonmay\}@usc.edu}}

\begin{document}
\maketitle

\begin{abstract}

Current Machine Translation (MT) models still struggle with more challenging input, such as noisy data and tail-end words and phrases.  Several works have addressed this robustness issue by identifying specific categories of noise and variation then tuning models to perform better on them.  An important yet under-studied category involves minor variations in nuance (non-typos) that preserve meaning w.r.t. the target language.  We introduce and formalize this category as \emph{Natural Asemantic Variation} (NAV) and investigate it in the context of MT robustness.  
We find that existing MT models fail when presented with NAV data, but we demonstrate strategies to improve performance on NAV by fine-tuning them with human-generated variations.  We also show that NAV robustness can be transferred across languages and find that synthetic perturbations can achieve some but not all of the benefits of organic NAV data.  


\end{abstract}

\section{Introduction}



While current Machine Translation (MT) models often perform decently in regular conditions, robustness to slight modifications on input remains a challenge.  Several works have investigated the devastating effects that minor noise can cause in MT \cite{belinkov2018synthetic, khayrallah-koehn-2018-impact}.   Addressing this is critical as robustness to naturally occurring variation is important in deployed systems.



Some classes of perturbations to which humans exhibit robustness but studies have shown MT models struggle with include speling errors or rtypos, CaSinG \cite{niu-etal-2020-evaluating}, v15sua11y-51m1ll4r ch4r4ct3r5 \cite{eger-etal-2019-text}, and synonym \sout{replacement} substitution \cite{wang-etal-2018-switchout}.  However, we focus on a less-known class we call Natural Asemantic Variation (NAV).



Whereas aforementioned classes tend towards perturbations that generate agrammatical, nonstandard, or unnatural sentences, NAV perturbations represent the extra-semantic linguistic properties of a language that allow for subtle changes in nuance while expressing the same core meaning.  Specifically, a NAV perturbation is a kind of paraphrase in a source language that would not affect its translation in a given target language.  Studying NAV is an important endeavor in MT robustness since it covers a less-contrived source of variation compared to existing MT robustness studies.  As this topic is under-studied, it has also not been well-defined.  We offer an example of NAV perturbations in Japanese in Table \ref{tab:nav} and will provide further definitions and examples throughout the paper to help solidify the reader's understanding.

\begin{table}[t]
    \centering
    \scalebox{0.67}{
    \begin{tabular}{c|c}
        Perturbation & Asemantic Variation \\
        \hline \\
        \begin{CJK}{UTF8}{min} 彼女は私に本を\underline{返した}　\end{CJK} & informal verb conjugation \\
        \hline \\
        \begin{CJK}{UTF8}{min} 彼女は\underline{俺}に本を返しました　\end{CJK} & masculine "me" pronoun \\
        \hline \\
        \begin{CJK}{UTF8}{min} 彼女は私に本を返し\underline{てくれ}ました　\end{CJK} & ``favor'' nuance \\
    \end{tabular}
    }
    \caption{\small Examples of NAV perturbations for a Japanese sentence, \begin{CJK}{UTF8}{min} 彼女は私に本を返しました　\end{CJK} (she returned the book to me).  These natural variations, which cause nuance differences in Japanese, cannot be translated succinctly into English.  Therefore, all perturbations can be translated reasonably to the same English sentence.  A robust MT model (or human) should be able to extract the same semantic meaning regardless of the variation provided.
    }
    \label{tab:nav}
    \vspace{-0.4cm}
\end{table}

From Table \ref{tab:nav}, we see how slight modifications of a sentence in one language can convey specific nuanced differences in that language, without warranting a change to its translation in another language.  A common challenge in MT has to do with the fact that there are almost always several correct answers, and we can imagine translationally-equivalent mappings between large sets of similar sentences in different languages.  Often we must resort to imperfect training and evaluating using subsets of these imagined sets due to normal data limitations.  However, leveraging the STAPLE dataset \cite{staple20}, we have a unique opportunity to investigate NAV phenomena more directly with access to several (often hundreds) of high-quality NAV perturbations.  These perturbations allow us to test for NAV robustness, evaluating model behavior on challenging NAV test sentences. 

In this paper, we contribute a formal definition for a linguistically-rich class of perturbations (NAV) and provide examples which help ground the concept (\S \ref{sec:def}).  We also contribute an evaluation setup to measure NAV robustness in the form of a repurposed test set from STAPLE and simple metrics to evaluate quality and consistency of MT models (\S \ref{sec:met}).  This formalization and metrics allow us to investigate specific research questions regarding NAV, as summarized below.   

We evaluate existing MT models for NAV robustness and characterize the observed errors related to underspecification resolution and rarer expressions (\S \ref{sec:nav}).  We show improvements in BLEU and our NAV-robustness metrics by exposing the model to NAV data and comparing various sub-selection and fine-tuning strategies (\S \ref{sec:trans}).  We even find that these improvements can be transferred across languages, improving NAV robustness in languages that have not seen any NAV data (\S \ref{sec:zero}).  We also design and perform synthetic perturbations, as these may provide a cheaper, more scalable way to improve NAV robustness without human annotations.  We do achieve improvements though organic data is still more useful than synthetic (\S \ref{sec:syn}).  Finally, we analyze our NAV-robust models in other evaluation settings with data that is out-of-domain and/or showing different classes of noise to investigate the side-effects of our methods (\S \ref{sec:add}). 



\section{NAV Robustness}


In robustness of NLP, it is difficult to determine improvements without considering both what types of input models are meant to be robust \emph{to} and what robustness \emph{looks like}. In this section, we will first discuss NAV with a formal definition and explanatory examples.  Then we will define NAV robustness in terms of desired behaviors and present metrics as proxies for measuring NAV robustness.

\subsection{NAV Perturbations}\label{sec:def}

To formally define \textit{natural asemantic variation}, we consider a corpus $C$ in two languages $\mathcal{X}, \mathcal{Y}$ that contain $c$ paired clusters of translationally-equivalent natural and grammatical sentences:
$$ C = \{(X_i, Y_i)\} \;\; \forall i \in \{1, ..., c\}; $$
$$ X_i = \{x_i^1, ..., x_i^n\}; \; Y_i = \{y_i^1, ..., y_i^m\}; $$
$$ \forall j \in \{1, ..., n\}, \; \forall k \in \{1, ..., m\}, \; x_i^j \Leftrightarrow y_i^k, $$
where $\Leftrightarrow$ represents \textit{translational equivalence}, meaning the two sentences $x_i^j$ and $y_i^k$, which are in clusters $X_i$ and $Y_i$, respectively, can be reasonably translated to each other. For example, all the Japanese examples in Table \ref{tab:nav} are translationally equivalent with \textit{she returned the book to me} in English. Another valid $y$ could be \textit{she gave the book back to me}.
All $x \in X_i$ can then be considered NAV$_{\mathcal{X},\mathcal{Y}}$ perturbations of each other. A NAV perturbation must be considered in context of the language pair.  $x_i^1$ and $x_i^2$ may be  NAV$_{\mathcal{X},\mathcal{Y}}$ perturbations of each other but not NAV$_{\mathcal{X},\mathcal{Z}}$ perturbations.\footnote{Note: A semantic paraphrase $x_i^1$ of a sentence $x_i^2$ in the same language $\mathcal{X}$ cannot be assumed to be a NAV perturbation \textit{on its own} because it may or may not cause a change to its translation in the relevant target language.}

In this work, we further limit the corpus to be:
$$ \forall i \;  |X_i| > 1 , \; |Y_i| = 1, $$
meaning only one reference translation is provided.  This choice is due to data limitations as we only have access to semantically-equivalent variations on the source side.  (if $|X_i| = 1$ as well, then $C$ has the form of most standard MT datasets).

\begin{figure}[t]
    \centering
    \includegraphics[scale=.4]{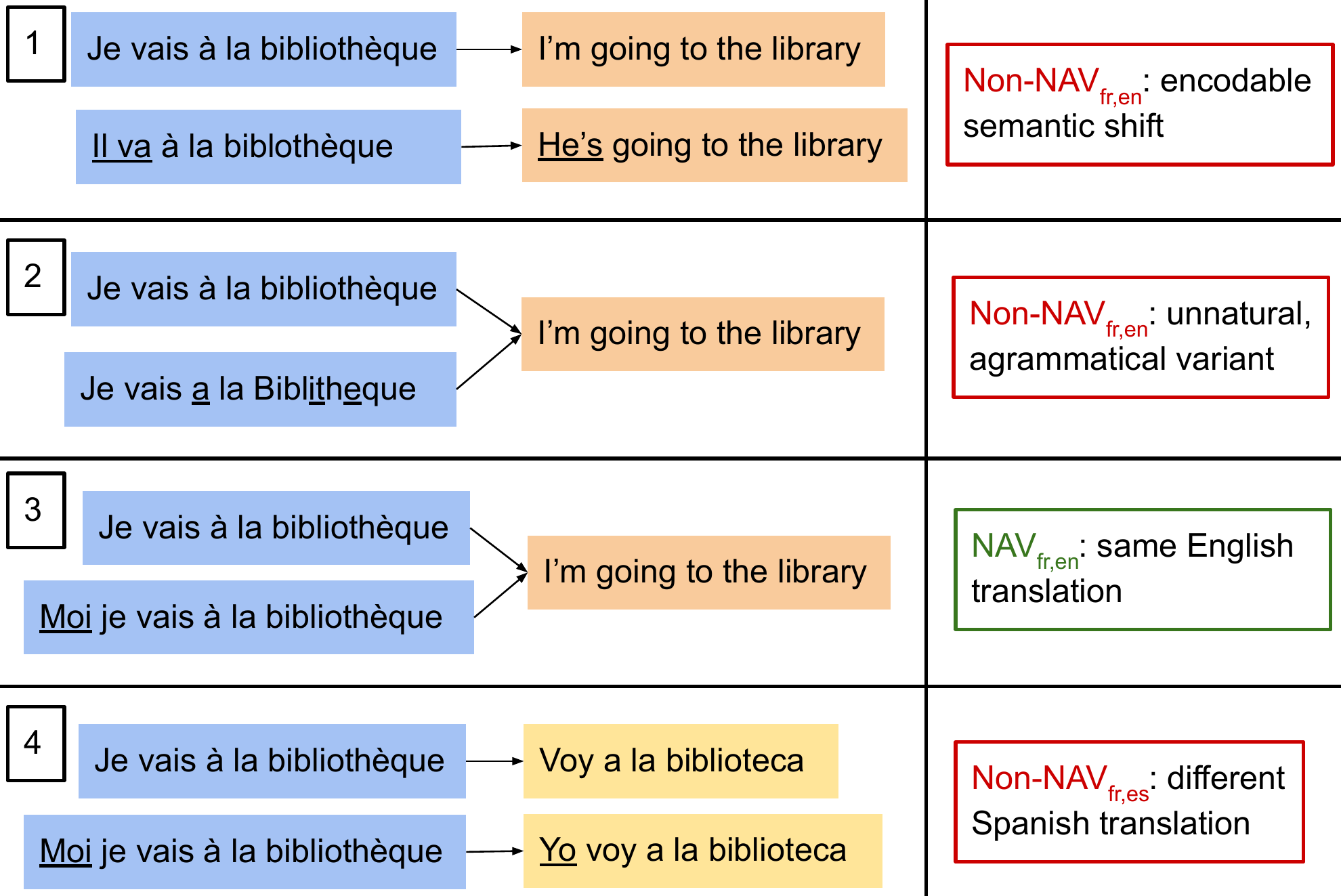}
    \caption{4 examples of perturbations to an original French sentence `Je vais à la bibliothèque'.  Only one perturbation (example 3) can be called a NAV perturbation as the variation is an emphasis on the subject, difficult to encode in written English.  The same perturbation is however not NAV in the context of a different target language that does easily encode subject emphasis (example 4).}
    \label{fig:NAV_examples}
    \vspace{-0.4cm}
\end{figure}

To ground these definitions, Figure \ref{fig:NAV_examples} offers specific examples to distinguish NAV perturbations from non-NAV perturbations.
The modifications in examples 1 and 2 are either nonstandard or have too large of an impact on semantics such that their translations are different.  Example 3 shows a simple example of a NAV perturbation while example 4 demonstrates the language dependence of NAV by using the same perturbation from 3, highlighting the complexity of this problem. For a different target language, the two sentences would no longer exist in the same $X_i$.

\subsection{NAV Robustness}\label{sec:met}
  
In order to evaluate models for their robustness to NAV perturbations, we consider two desiderata, \textit{quality} and \textit{consistency}, which we evaluate separately.
There are often several high-quality answer options in MT, so two models could both produce high-quality translations but differ greatly in their consistency (how similar their hypotheses among minimally-altered, translationally-equivalent inputs are).  For our purposes, we posit that a more NAV-robust MT model should not only have increased or maintained translation quality but also consistency in output when given different NAV perturbations as input.  Consistency in output would help ensure overall system robustness if we imagine the MT system being part of a larger NLP pipeline.

In this work, for translation quality, we use BLEU \cite{papineni-etal-2002-bleu} with SacreBLEU~\footnote{\texttt{nrefs:1|case:lc|eff:no|tok:13a| smooth:exp|version:2.0.0}} \cite{post-2018-call}.  For translation consistency we define a metric, CONSIST$^{\theta}$, which rewards a translation model $f_\theta$ for exhibiting less variation in its outputs $\hat{Y}^{\theta}_i=f_\theta(X_i)$ among NAV-perturbed inputs $X_i$.  Specifically, for a pair $(X_i,\hat{Y}^{\theta}_i)$, CONSIST$^{\theta}$ is calculated as follows:
$$ CONSIST^{\theta}_i \triangleq \frac{1}{|X_i|}\sum_{j=1}^{|\hat{Y}^{\theta}_i|}\frac{|\hat{y}_i^j|}{j}, $$
\noindent where $\hat{Y}^{\theta}_i \triangleq \{\hat{y}_i : \exists x^j_i \in X_i \wedge \theta(x^j_i) = \hat{y}_i\}$, 
$|\hat{y}_i| \triangleq |\{x_i : x_i \in X_i \wedge \theta(x_i) = \hat{y}_i\}|$, 
and we order  $\hat{Y}^{\theta}_i$ as 
$[\hat{y}_i^1, ..., \hat{y}_i^m]$, sorted descending by $|\hat{y}_i^j|$. 
Intuitively, $|\hat{y}_i^j|$ is the number of $x_i \in X_i$ for which $\theta(x_i)$ outputs hypothesis $\hat{y}_i^j$.  
We then average this score across all translation cluster-pairs in the corpus:
$$ CONSIST^{\theta} \triangleq \frac{1}{c} \sum_{i=1}^{c} CONSIST^{\theta}_i.$$
We drop the superscript when unambiguous. This metric may break down in long-output tasks but works well for our domain of short, simple sentences.  Essentially, this metric works by punishing the model more and more for novel translation outputs on translationally-equivalent inputs.  An explanatory example calculation is available in Appendix \S \ref{sec:consist}.  We also discuss an alternative consistency metric with trade-offs in Appendix \S \ref{sec:alt}.

\section{Analysis Setup}\label{sec:anal}

In this work, we seek to provide answers to several analysis questions related to NAV perturbations and MT models' robustness to them.  Using our quality and consistency metrics we investigate the following:

\begin{itemize}
    \item Are existing MT models robust to NAV? if not, what types of errors are the models making?
    \item How can NAV perturbation data be used to improve an MT model's NAV robustness? 
    \item Can a model's improved NAV robustness in one language pair be transferred to another language pair?
    \item How do synthetic perturbations compare to organic NAV data w.r.t. NAV robustness?
    \item What behavior changes does a `NAV-robustified' MT model exhibit in other MT contexts such as out of domain data or on other types of robustness tests?
\end{itemize}

\subsection{Baseline Models}\label{sec:base}

We work with two main classes of baseline models: \textit{mono-pair} (can translate one specific language to one specific other language, e.g. ja-en) and \textit{multi-pair} translation models (can translate between several languages, e.g. \{hu,ja,pt\}-\{hu,ja,pt\}).  All models are transformer models \cite{NIPS2017_3f5ee243} using the fairseq toolkit \cite{ott-etal-2019-fairseq}.  Our principal models are multi-pair since many of our analysis questions relate to language transfer, but we also perform preliminary experiments with mono-pair models for completeness.  

We pre-train mono-pair models with the \href{https://tatoeba.org/en/}{Tatoeba} corpus for \{hu,ja,pt\}-en.  This corpus is fairly close to the STAPLE domain and allows for reasonable baseline models.  We obtain subwords and tokenize using SentencePiece \cite{kudo-richardson-2018-sentencepiece}.  Hyperparameters are the default settings for the \texttt{transformer} architecture in fairseq.

For multi-pair model experiments, we use the M2M-100 model \cite{JMLR:v22:20-1307}, which is trained on CC-Matrix \cite{schwenk-etal-2021-ccmatrix} and CC-Aligned \cite{el-kishky-etal-2020-ccaligned}.  This model comes with a predetermined vocabulary and tokenizer.  It uses the \texttt{transformer\_wmt\_en\_de\_big} architecture in fairseq with 24 encoder and 24 decoder layers.  The checkpoints available online are used as initialization for further fine-tuning using different NAV data, as described next.\footnote{More details on all the specifications for our models are available in experiment scripts released publicly in our \href{https://github.com/jlbrem/nav_robustness}{github}}.


\begin{figure}[t]
    \centering
    \includegraphics[scale=.33]{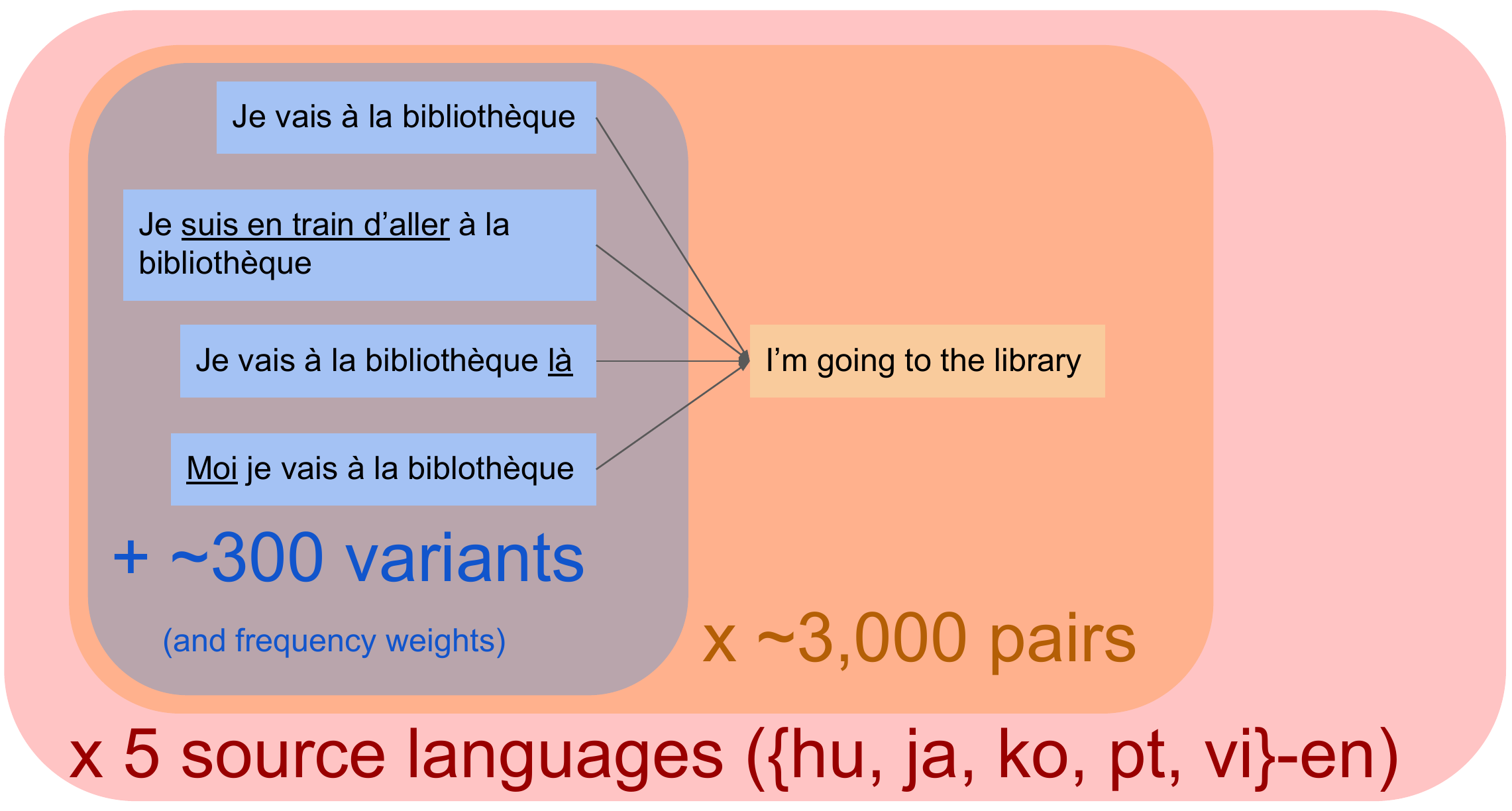}
    \caption{Diagram describing the STAPLE data.  French is used to mirror the French examples used earlier for ease of relating NAV concepts.}
    \label{fig:staple}
    \vspace{-0.4cm}
\end{figure}

\subsection{NAV Data}\label{sec:data}

This research is made possible largely by the unique dataset publicly released by Duolingo for the STAPLE shared task \cite{staple20}. For each English sentence, several (average ~300) accepted translations of that sentence in one of five languages were sourced from Duolingo users and further annotated with frequency scores based on how often they used that specific translation. We use the enumerated valid translations as source-side NAV perturbations translating to the same target English sentence.  A diagram representing the data can be seen in Figure \ref{fig:staple}.  Table \ref{tab:nav} includes a real example from the training split for Japanese-English. 

For NAV fine-tuning (\S \ref{sec:nav}), we use different subsets of the STAPLE training corpus in three language pairs \{hu,ja,pt\}-en.  For each ``many-to-one'' pair, we either use \textbf{all} of the many NAV perturbations or subselect.  When subselecting, we consider three strategies.  Since we also have frequency scores for each perturbation, we could select a number of the \textbf{most} frequent perturbations, \textbf{least} frequent perturbations, or uniformly sample for a \textbf{random} subset.  In our paper, a subset is defined by a number and a strategy, so ``10-random'' means 10 random NAV perturbations were chosen for each example.

For NAV robustness evaluation, we use the full (all NAV perturbations included) many-to-1 pairs of the STAPLE test split.  On top of the \{hu,ja,pt\}-en sets, we hold out \{ko,vi\}-en test sets to evaluate 0-shot language transfer using multi-pair models (\S \ref{sec:trans}). Data statistics are shown in Table \ref{tab:stats}.

\subsection{Synthetic Data Augmentation}\label{sec:synth}

Obtaining NAV perturbations is expensive, requiring human annotation by bilingual language experts to generate natural variations that don't affect translation equivalence.  We consider applying synthetic perturbations to some of the STAPLE data to compare potential NAV robustness gains with a cheaper, more scalable strategy.

\begin{table}[t]
    \centering
    \begin{tabular}{c||c|c|c}
        lang & train & dev & test \\
        \hline
        hu & 250k/4k & 28k/500 & 34k/500 \\
        ja & 860k/2.5k & 170k/500 & 170k/500 \\
        pt & 530k/4k & 60k/500 & 68k/500 \\
        ko & -/- & -/- & 150k/500 \\
        vi & -/- & -/- & 28k/500
    \end{tabular}
    \caption{Statistics for STAPLE data used in our experiments. \emph{unique source segments} / \emph{unique target segments}}
    \label{tab:stats}
    \vspace{-0.4cm}
\end{table}

With the organic, human-generated variations in the STAPLE data, we create a synthetic dataset by taking the 1-most frequent source sentence from each translation pair and perturbing it synthetically 9 times for a total of 10 variations (while the 10 copies of the target sentence are left unchanged).

For our roman script languages \{hu,pt\}-en, we use known noising techniques of random casing changes and character substitutions, insertions and deletions as in \citet{niu-etal-2020-evaluating}.  These perturbations are not NAV perturbations, but they represent a common and easy way to add noisy data.

We also add synthetic NAV perturbations for the ja-en split.  Roman script noising strategies are not perfectly translatable to ja-en.  Also, ja-en arguably has the easiest rule-based modifications that can be programmed to automatically generate NAV perturbations.  By no means are these methods exhaustive, but a combination of simple insertions of emphasis-related particles, substitutions of pronouns based on gender identity, and dropping unnecessary pronouns serve as a basic technique for synthetic NAV noising.  Rules were implemented after scanning training data and observing patterns of NAV perturbation.\footnote{Perturbation scripts are available on \href{https://github.com/jlbrem/nav_robustness}{github}.  Rules are described in Appendix \S \ref{sec:rules}.} 

\subsection{Additional Evaluations}

We use other existing MT datasets to evaluate our models in contexts such as performance on out-of-domain (OOD) data (\S \ref{sec:ood}) and robustness-transfer to other types of noisy input (\S \ref{sec:rob}).  We use test splits from Tatoeba, OPUS-100 \cite{zhang-etal-2020-improving} and MTNT \cite{michel-neubig-2018-mtnt}.

  We also create additional evaluation sets from the STAPLE data to analyze in-domain model behavior. In addition to the `all perturbations' test sets, we also filter out `1-most' and `1-least' test sets.  These provide a more standard MT evaluation framework that relies on BLEU to measure in-domain translation performance on common (`1-most') and robustness to uncommon, NAV-perturbed (`1-least') sentence pairs.

\section{Experiments and Results}

For our experiments and results section, we will progress chronologically as we address each of our analysis questions as defined in Section \ref{sec:anal}.



\subsection{Existing MT NAV Behavior}\label{sec:nav}


Before we perform comprehensive experiments to analyze NAV robustness, we investigate how existing MT models perform on NAV perturbations.  This way we can establish 1) if this even is a problem in MT and 2) if so, what types of errors are we aiming to address?  We simply take our baseline models (\S \ref{sec:base}) and search for error patterns produced from example inputs in the STAPLE dataset.

One pattern of errors involves an inability to properly resolve cases of under-specification.  For example, the M2M-100 model properly handles ``\begin{CJK}{UTF8}{min} \textbf{それは}トマトの種類です。\end{CJK}'', outputting ``This is a type of tomato.''  However, a NAV-perturbed input, ``\begin{CJK}{UTF8}{min} トマトの種類です。\end{CJK}'' causes the model to output `Species of tomatoes.' since this sentence drops the subject, common in Japanese.

Another pattern of errors involves an inability to handle less common synonyms or paraphrases for the same target word(s).  `Please' is the main way to express politeness in a request in English but other languages have multiple common ways to express this.  The M2M-model properly translates the Portuguese sentence ``por \textbf{favor}, não a desperte agora.'' as ``Please don’t wake her up now.'' but mistakenly translates ``por \textbf{obséquio}, não a desperte agora.'' as ``For obsession, don’t wake her up now.''.

In just a small sample of examples, we find several instances of these types of errors, confirming our intuition that existing MT models cannot handle NAV perturbations completely.  We are also able to see that these types of mistakes will be captured by our BLEU and CONSIST metrics, motivating our quantitative experiments.  More information on this error analysis is available in Appendix \S \ref{sec:quant}.

\subsection{Improving NAV Robustness with NAV Data}\label{sec:trans}

For our first experiments, we fine-tune our mono-pair transformer models pre-trained on Tatoeba using strategies for subselecting STAPLE data discussed in Section \ref{sec:data} and evaluate using metrics from Section \ref{sec:met}.  After initially experimenting with several combinations of number of perturbations per set pair and selection strategy,\footnote{full results in appendix} we report on 4 conditions, representative of common ``real-world'' MT scenarios:\footnote{We choose to only report on these conditions to strike a balance between not overloading the reader with too many experiment settings and providing enough that useful conclusions can be drawn by comparing.}


\textbf{baseline:} off-the-shelf MT model, no fine-tuning (\S \ref{sec:base})

\textbf{+ 1-most:} simple domain adaptation using parallel text in target domain (one ``typical'' translation per sentence).

\textbf{+ 10-random:} preferred ``NAV robustness'' data condition, having a small yet diverse set of NAV perturbations per translation pair

\textbf{+ all:} ``throw in all the data'' approach

\begin{figure}[h]
    \centering
    \includegraphics[scale=.7]{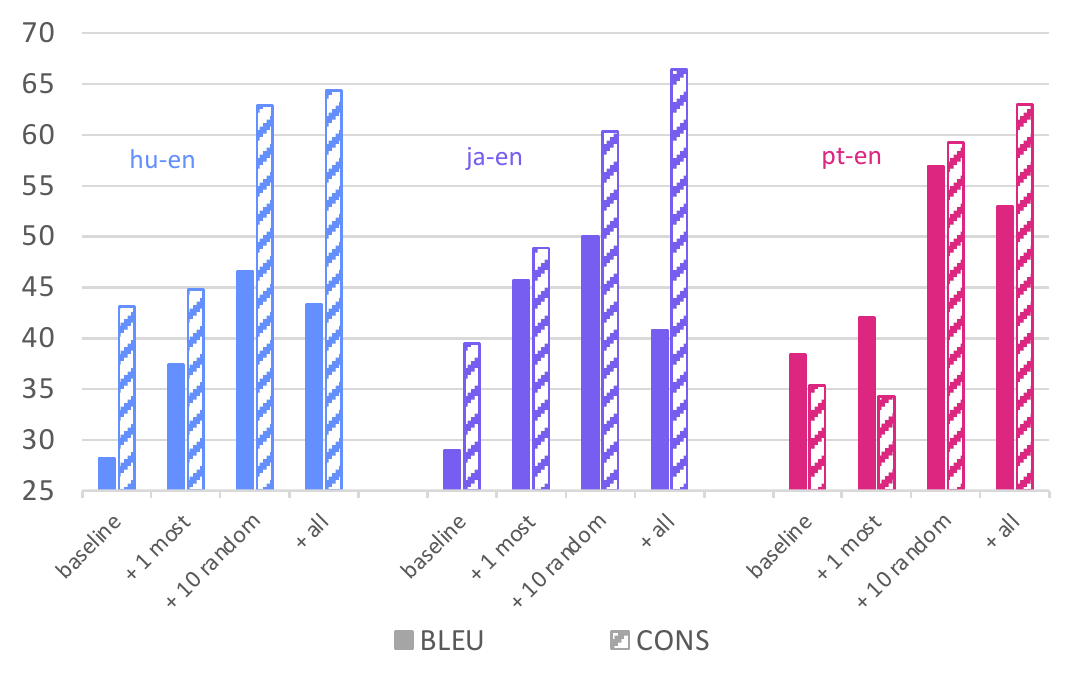}
    \caption{Results from mono-pair model experiments. The `+ 1 most' condition shows improvements over a baseline in a typical MT domain adaptation scenario where some target domain data is available.  The `+ 10 random' condition shows how exposing the model to NAV perturbations further increases both BLEU and CONSIST scores beyond domain adaptation with frequent input forms.  While including all NAV perturbations (`+ all') continues to improve CONSIST, we see a decrease in BLEU.}
    \label{fig:DIR}
    \vspace{-0.4cm}
\end{figure}

Our results from these experiments are shown in Figure \ref{fig:DIR}.  The main purpose of these preliminary experiments is to provide evidence that NAV robustness improvement is possible on smaller scale models.  We test various configurations to justify design choices for our future, more comprehensive experiments with large multi-pair models.  We find that the `+ 10-random' condition results in the best trade-off of BLEU and CONSIST.

 We repeat these experiments using a large multi-pair model.  We find similar patterns in the results, confirming ``10-random'' as a decent strategy for improving NAV robustness.  Here, the baseline is the M2M-100 model and models are fine-tuned and evaluated on only the same language pair.\footnote{To help validate our results, we run 3 different seeds for the randomization of data in `random-10.'  Standard deviation for BLEU scores is between 0.4 and 0.6.  Standard deviation for CONSIST scores is between 0.7 and 1.  The variation is small enough to not affect conclusions.} As the patterns are similar, results are plotted in the Appendix \ref{sec:multi}.

\subsection{Transfering NAV Robustness across Languages}\label{sec:zero}

\begin{figure}[h]
    \centering
    \includegraphics[scale=.7]{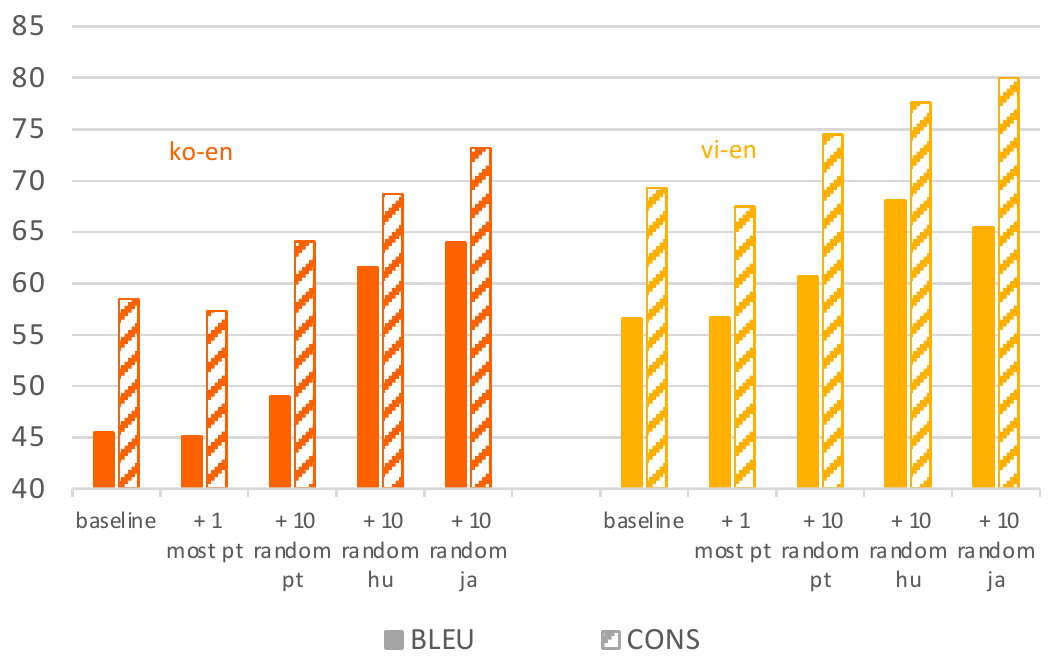}
    \caption{As in mono-pair experiments, including 10 NAV perturbations per pair shows larger robustness improvement than simply fine-tuning on 1-most.}
    \label{fig:TRANS}
    \vspace{-0.2cm}
\end{figure}

 With a large multi-pair model, we can consider zero-shot language-transfer of NAV robustness.  First, we take the M2M-100 baseline models and fine-tune them on one of the STAPLE fine-tuning language-pairs.  We then evaluate on our held-out language test sets (unseen pairs during fine-tuning, \{ko,vi\}-en).  Our results are shown in Figure \ref{fig:TRANS}. From previous experiments, we see sufficient evidence that `` + 10 random '' is appropriate for NAV robustifying, so we mainly report on this setting.

Results suggest zero-shot transfer of NAV robustness is possible, with 10-random NAV perturbations per pair showing larger robustness improvement than simply fine-tuning on 1-most. We also observe language differences. ko-en BLEU improves more from ja-en fine-tuning while vi-en improves more from hu-en.

\begin{figure}[h]
    \centering
    \includegraphics[scale=.7]{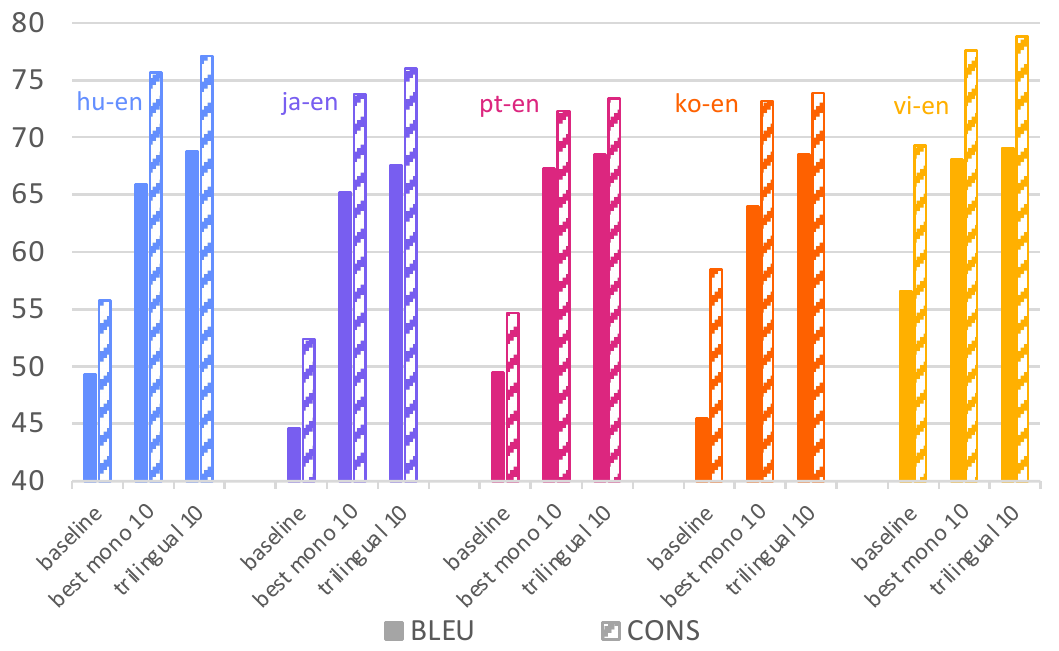}
    \caption{Multilingual (trilingual) fine-tuning improves both BLEU and CONSIST for every evaluation language pair compared to that pair's previous best model using monolingual fine-tuning. }
    \label{fig:TRI}
\end{figure}

Finally, we perform multilingual fine-tuning experiments in which we combine all three training sets together \{hu,ja,pt\}-en with the `+ 10 random' strategy and evaluate on all 5 test sets.  We compare the multilingual fine-tuning results to the previous best results (according to BLEU) from our monolingual fine-tuning experiments.  Results are shown in Figure \ref{fig:TRI}.  We see that using all 3 test sets for NAV fine-tuning is consistently a better option over any given single set.

\subsection{Approaching NAV Robustness with Synthetic Data Augmentation}\label{sec:syn}

\begin{figure}[h]
    \centering
    \includegraphics[scale=.6]{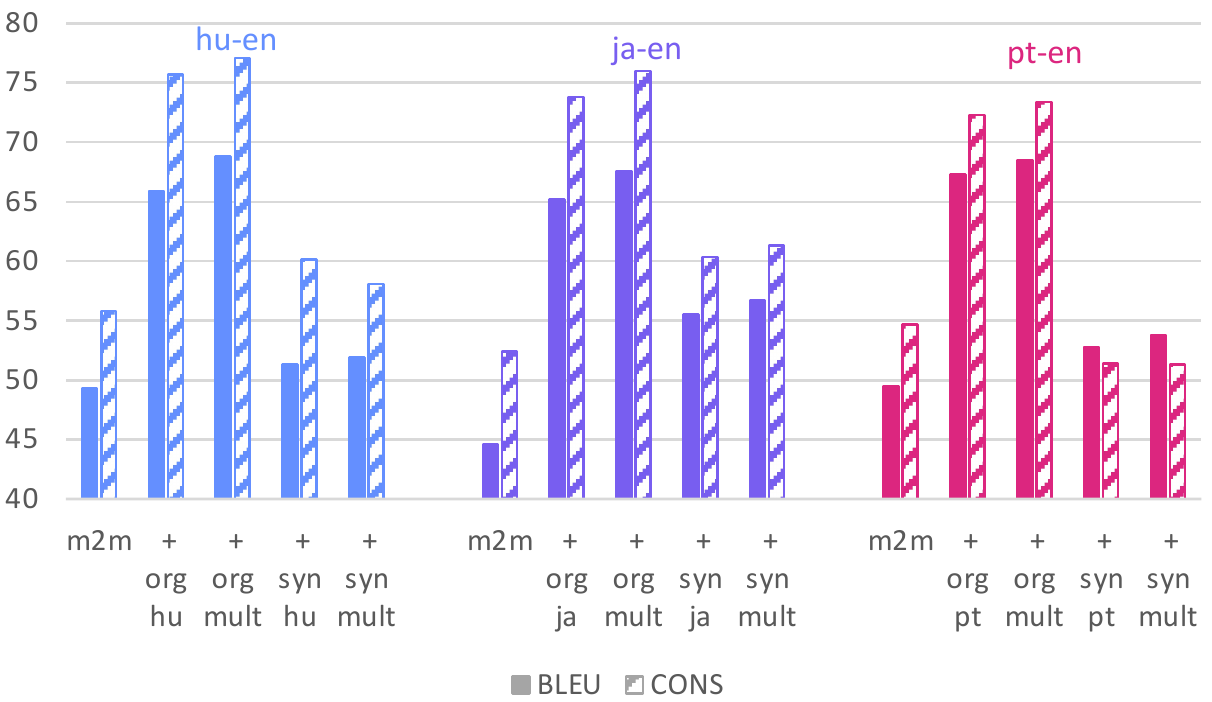}
    \caption{BLEU and CONSIST scores for organic vs. synthetic perturbations during fine-tuning.  Synthetic perturbations improve robustness over baselines but organic perturbations are much more useful.}
    \label{fig:syn}
\end{figure}

We repeat experiments from the previous section using our synthetic fine-tuning sets (Section \ref{sec:synth}) and compare to the organic 10-random sets.  We also combine all the synthetic sets \{hu,ja,pt\}-en for multilingual fine-tuning and evaluate on our NAV robustness evaluation sets.  Results are shown in Figure \ref{fig:syn}.  

Our synthetic fine-tuning sets do improve robustness compared to a baseline model, but they are far from achieving the same improvements as the organically generated data.  We also see how our NAV-oriented synthetic data (ja-en) more closely approximates organic NAV data gains compared to our non-NAV synthetic data (\{hu,pt\}-en).

\begin{figure}[h]
    \centering
    \includegraphics[scale=.6]{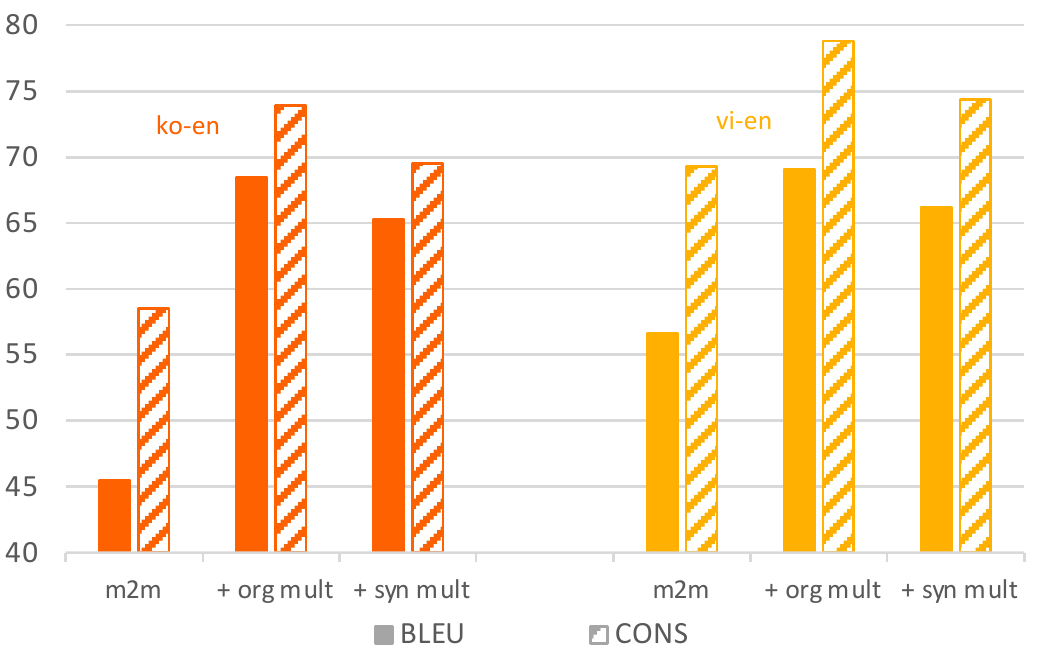}
    \caption{BLEU and CONSIST scores for organic vs. synthetic perturbations during fine-tuning in zero-shot language transfer.  Synthetic perturbations improve robustness over baselines but organic perturbations are much more useful. However, this gap is smaller than that shown with in-language experiments. }
    \label{fig:synmult}
    \vspace{-0.4cm}
\end{figure}

We also test synthetic fine-tuning performance in zero-shot language transfer. Results are shown in Figure \ref{fig:synmult}.  We continue to see that synthetic data is not able to provide the same NAV robustness gains as organically-generated data.  However, in zero-shot language transfer, multilingual synthetic data more closely approaches organic-data performance than it can with in-language fine-tuning.  This suggests in-language organic NAV data is most useful even if some NAV robustness can be improved with transfer.

\subsection{Additional Evaluations}\label{sec:add}

Results from the previous sections suggest that NAV robustness can be effectively learned (even in zero-shot scenarios) by exposing models to a variety of several NAV perturbations per translation example.  However, these findings raise new questions about what specifically the models are learning, how transferable that learning is and how else the model's behavior changes.

We perform several analysis experiments in which we take our best models from previous experiments, which we designate as our `NAV-robust' models and evaluate them compared to baseline MT models in other evaluation settings.  These new test settings include OOD data (\S \ref{sec:ood}), other classes of noise (\S \ref{sec:rob}) and more frequent (less NAV-relevant) in-domain data (\S \ref{sec:in}). 

The overall takeaway from these experiments is that our NAV-robust models sacrifice some performance in non-NAV-relevant settings.  This is not uncommon in general NLP fine-tuning, as often models fine-tuned for a new task, domain, etc. often underperforms in the original setting \cite{thompson-etal-2019-overcoming,he-etal-2021-analyzing}.  

These experiments serve primarily to confirm that there are limitations and drawbacks to improving NAV robustness with our current proposed methods.  We include specifics in the Appendix to distill the main work for novel contributions.  These results suggest avenues for future work to avoid these trade-offs for robustness, which remains a problem in the wider NLP community.

One set of results we communicate here is performance on in-domain `1-most' and `1-least' test data.  These test settings are useful because they allow us to directly compare models on similar text examples that differ in frequency of use, thus distinguishing between `normal' sentences and heavily-NAV-perturbed sentences.  The latter evaluation serves as another proxy for robustness.

\begin{figure}[h]
    \centering
    \includegraphics[scale=.6]{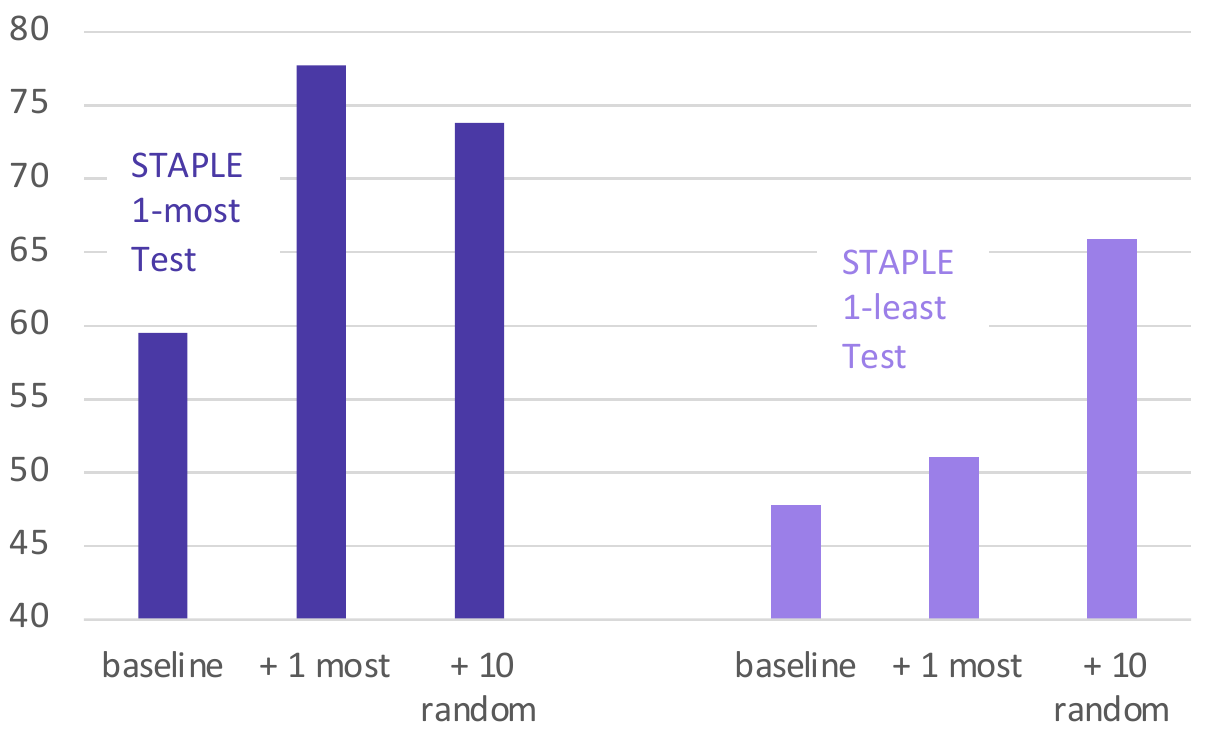}
    \caption{BLEU scores evaluating on 1-most and 1-least STAPLE test sets in ja-en.  Fine-tuning with common `1-most' data is better than using NAV perturbations for testing on 1-most inputs.  However, testing on rarer inputs exhibiting NAV phenomena in the 1-least test data, intentional NAV robustifying (fine-tuning with `+ 10 random' data) is necessary over the common 1-most data.}
    \label{fig:mostleast}
\end{figure}

In most of our NAV experiments from earlier sections, `+ 10 random' consistently performs better than `+ 1 most'.  This is reflected in the `1-least' evaluation as well.  However, in the `1-most' evaluation, `+ 1 most' outperforms `+ 10 random'.  What this shows is that evaluating on frequent, non-noisy data may select for models that actually perform worse on data of the same domain but with less frequent NAV forms.  These results can be seen for ja-en in Figure \ref{fig:mostleast} and full results (which mirror those shown here) are shown in \S \ref{sec:in}.

\begin{table}[t]
\hspace{-0.2cm}
\scalebox{0.53}{
\begin{tabular}{l||cc|rr|rr||cc|cl}
\multicolumn{1}{c||}{} & \multicolumn{2}{c|}{hu}                                                                 & \multicolumn{2}{c|}{ja}                                                                          & \multicolumn{2}{c||}{pt}                                                                          & \multicolumn{2}{c|}{ko}                                 & \multicolumn{2}{c}{vi}                                                                 \\ \hline \hline
Tatoeba Transformer   & {\color[HTML]{3531FF} 28.2}                           & 43.1                            & {\color[HTML]{3531FF} 29.0}                                    & 39.5                            & {\color[HTML]{3531FF} 38.4}                                    & 35.31                           & {\color[HTML]{3531FF} -}                   & -          & {\color[HTML]{3531FF} -}                              & -                              \\
+ 1 most              & {\color[HTML]{3531FF} 37.4}                           & 44.8                            & {\color[HTML]{3531FF} 45.7}                                    & 48.9                            & {\color[HTML]{3531FF} 42.0}                                    & 34.24                           & {\color[HTML]{3531FF} -}                   & -          & {\color[HTML]{3531FF} -}                              & -                              \\
+ 10 random           & {\color[HTML]{3531FF} {\ul 46.6}}                     & 62.9                            & {\color[HTML]{3531FF} {\ul 50.0}}                              & 60.3                            & {\color[HTML]{3531FF} {\ul 56.9}}                              & 59.22                           & {\color[HTML]{3531FF} -}                   & -          & {\color[HTML]{3531FF} -}                              & -                              \\
+ all                 & {\color[HTML]{3531FF} 43.3}                           & {\ul 64.4}                      & {\color[HTML]{3531FF} 40.8}                                    & {\ul 66.4}                      & {\color[HTML]{3531FF} 53.0}                                    & {\ul 62.95}                     & {\color[HTML]{3531FF} -}                   & -          & {\color[HTML]{3531FF} -}                              & -                              \\ \hline \hline
M2M-100               & \multicolumn{1}{r}{{\color[HTML]{3531FF} 49.3}}       & \multicolumn{1}{r|}{55.8}       & {\color[HTML]{3531FF} 44.6}                                    & 52.4                            & {\color[HTML]{3531FF} 49.5}                                    & 54.7                            & {\color[HTML]{3531FF} 45.5}                & 58.5       & \multicolumn{1}{r}{{\color[HTML]{3531FF} 56.6}}       & \multicolumn{1}{r}{69.3}       \\
+ 1 most              & \multicolumn{1}{r}{{\color[HTML]{3531FF} 52.7}}       & \multicolumn{1}{r|}{60.6}       & {\color[HTML]{3531FF} 55.8}                                    & 60.0                            & {\color[HTML]{3531FF} 51.2}                                    & 47.3                            & {\color[HTML]{3531FF} 59.4}                & 66.1       & \multicolumn{1}{r}{{\color[HTML]{3531FF} 64.5}}       & \multicolumn{1}{r}{71.8}       \\
+ 10 random           & \multicolumn{1}{r}{{\color[HTML]{3531FF} {\ul 65.9}}} & \multicolumn{1}{r|}{75.7}       & {\color[HTML]{3531FF} {\ul 65.2}}                              & 73.8                            & {\color[HTML]{3531FF} {\ul 67.3}}                              & 72.3                            & {\color[HTML]{3531FF} {\ul 64.0}}            & 73.2       & \multicolumn{1}{r}{{\color[HTML]{3531FF} {\ul 68.1}}} & \multicolumn{1}{r}{77.6}       \\
+ all                 & \multicolumn{1}{r}{{\color[HTML]{3531FF} 65.5}}       & \multicolumn{1}{r|}{{\ul \textbf{78.8}}} & {\color[HTML]{3531FF} 55.3}                                    & {\ul 75.2}                      & {\color[HTML]{3531FF} 66.7}                                    & {\ul \textbf{77.6}}                      & {\color[HTML]{3531FF} 59.0}                  & {\ul \textbf{74.8}} & \multicolumn{1}{r}{{\color[HTML]{3531FF} 66.0}}         & \multicolumn{1}{r}{{\ul \textbf{78.8}}} \\ \hline
+ multi organic       & {\color[HTML]{3531FF} {\ul \textbf{68.8}}}            & {\ul 77.1}                      & \multicolumn{1}{c}{{\color[HTML]{3531FF} {\ul \textbf{67.6}}}} & \multicolumn{1}{c|}{{\ul \textbf{76.0}}} & \multicolumn{1}{c}{{\color[HTML]{3531FF} {\ul \textbf{68.5}}}} & \multicolumn{1}{c||}{{\ul 73.4}} & {\color[HTML]{3531FF} {\ul \textbf{68.5}}} & {\ul 73.9} & {\color[HTML]{3531FF} {\ul \textbf{69.1}}}            & {\ul \textbf{78.8}}                     \\
+ synthetic           & {\color[HTML]{3531FF} 51.3}                           & 60.1                            & \multicolumn{1}{c}{{\color[HTML]{3531FF} 55.5}}                & \multicolumn{1}{c|}{60.3}       & \multicolumn{1}{c}{{\color[HTML]{3531FF} 52.8}}                & \multicolumn{1}{c||}{51.4}       & {\color[HTML]{3531FF} 60.1}                & 65.8       & {\color[HTML]{3531FF} 64.5}                           & 74.8                           \\
+ multi synthetic     & {\color[HTML]{3531FF} 51.9}                           & 58.1                            & \multicolumn{1}{c}{{\color[HTML]{3531FF} 56.7}}                & \multicolumn{1}{c|}{61.3}       & \multicolumn{1}{c}{{\color[HTML]{3531FF} 53.8}}                & \multicolumn{1}{c||}{51.3}       & {\color[HTML]{3531FF} 65.3}                & 69.5       & {\color[HTML]{3531FF} 66.2}                           & 74.4                          
\end{tabular}
}
\caption{Main results from all NAV experiments organized by source language (columns) and baseline/fine-tuning method (rows).  In each column, the left {\color[HTML]{3531FF} blue} elements are BLEU scores, the right black elements are CONSIST scores.}
\label{tab:all}
\vspace{-0.4cm}
\end{table}

\section{Discussion}

Overall our work raises several questions about NAV phenomena and how to address them in MT.  We are able to provide some answers to a few research questions and open pathways for future work.  NAV describes a subtle yet substantial mark of language that stands as an appropriate target for current NLP research.

Addressing NAV in a systematic way is an incredible challenge due to the vastness of possible variation.  The STAPLE dataset provides a suitable way to begin investigating solutions to NAV-related problems in MT.  We are able to show how NAV robustness can be improved using STAPLE data and how those improvements can transfer across languages.  There seems to be language dependence whereby robustness learned in one language can have varying effects depending on the language transferred to.

We find that NAV-robust models can perform worse in non-NAV settings, which does not differ from several other findings in robustness work in which a more robust model may not perform better on original, non-noised input.  We also address scaling issues by synthetically generating NAV and non-NAV examples.  The synthetic NAV examples seem to help more than the non-NAV examples, but none of them are able to obtain the same improvements as human-created NAV data.

The major results from all of our experiments are displayed in Table \ref{tab:all} for side-by-side comparison.\footnote{This table offers a concise, overarching view of this work's experiments, as contrasted with the extracted figures in Section 4, which serve to highlight evidence pertaining to specific research questions.}

\section{Related Work}


\paragraph{Paraphrasing} Paraphrasing has been shown to be beneficial in MT contexts.  \citet{Hu_Rudinger_Post_Van-Durme_2019} have done work in releasing and using paraphrase data.  \citet{khayrallah-etal-2020-simulated} use generated paraphrases as data augmentation for MT.  One thing that differentiates our work is the formal definition of NAV.  It can be difficult to clarify what is or is not a paraphrase, and this work attempts to formalize a specific kind of paraphrase.

\paragraph{MT Robustness} Several works investigate robustness in MT by considering different classes of perturbations and developing strategies to improve performance at test time. \citet{niu-etal-2020-evaluating} perform misspelling- and casing-related perturbations on MT test input and evaluate different models on their robustness to these. \citet{salesky-etal-2021-robust} address several classes of perturbations by replacing a standard unicode-based encoder with a visual encoder, which demonstrates higher robustness to perturbations associated with visual appearance of language, such as 1337speak. \citet{zhang-etal-2020-improving} generate additional clean and noisy data using back-translation and different datasets to improve performance on noisy text.  We add to this line of work by addressing a specific kind of robustness (NAV).

\paragraph{Data Augmentation} There has also been substantial work in using synthetic data to improve MT more generally.  \citet{sennrich-etal-2016-improving} use back translation to augment MT datasets, which improves performance.  \citet{wang-etal-2018-switchout} augment data by reusing sentences with synonyms replaced.  \citet{karpukhin-etal-2019-training} and \citet{berard-etal-2019-machine} improve robustness by adding synthetic noise.  Our work is unique in our attempts to augment data using developed rules based on asemantic under-specification in language.



\section{Conclusion}

We present the reader with an under-studied subject in MT and contribute more formalized definitions and simple evaluation metrics for the phenomenon of natural asemantic variation.  We also perform experiments, showing how NAV perturbations can be used during fine-tuning to improve robustness of MT models, even when evaluating on a different language-pair.  Several questions remain to be explored to further formalize NAV, investigate its role in NLP modeling and improve techniques to increase nuanced understanding.

\section{Limitations}

Several of the limitations of this work are discussed throughout the paper and in the Appendix.  To summarize, the metrics used to evaluate NAV robustness may not work as well in other domains such as data with longer sentences.  However, the introduced formalism of NAV is relevant to any domain.

A limitation to mention about the formalism is that it likely does not perfectly distinguish NAV and non-NAV sentences.  Just as translators can argue about translation decisions, there are arguments for why or why not certain sentences should be considered `translationally-equivalent'.  `Asemantic' is also a term that has gray areas.  Nonetheless, the formalism allows us to establish metrics and a foundation for analysis for this very real phenomenon.

Our methods are also limited since they rely on access to unique high-quality data.  We address this with language-transfer and synthetic perturbations, but we are unable to obtain the same benefits as human-generated in-language data.

Other limitations include language coverage.  We work with 6 languages, which offers more support for generalization than only one language pair but still could raise some doubts about applicability to all languages.

There are also equipment limitations to consider.  Most of our experiments use the M2M-100 model as a baseline, which required our experiments to be run on RTX 6000 GPUs.  We do run some of the experiments on smaller baseline models which fit on older RTX 2800 machines and see similar results.

\section*{Acknowledgments}
This material is based upon work supported by the Defense Advanced Research Projects Agency (DARPA) under Agreement No. HR00112290056. Xiang Ren’s research is supported in part by the Office of the Director of National Intelligence (ODNI), Intelligence Advanced Research Projects Activity (IARPA), via Contract No. 2019-19051600007 and NSF IIS 2048211.

\bibliography{anthology,custom}
\bibliographystyle{acl_natbib}

\clearpage
\appendix

\section{Appendix}

\subsection{CONSIST Metric Example}\label{sec:consist}

Say, as an example, we have 1000 NAV perturbations of the same source sentence.  The model will then have 1000 output sentences, many of which will (hopefully) be the same.  We can then create groups of identical translation outputs and rank them by size from largest to smallest.  In our example, the model outputs 3 different translations, one 750 times, one 200 times, and one 50 times. We define $CONSIST$ as:

$$ CONSIST = \frac{1}{n} \sum_{i=1}^n \frac{|G|_i}{i} $$

\noindent where $i$ is the rank of the group, $n$ is the number of total input-output example pairs and $|G|_i$ is the ratio of the size of the $i$th largest group to $n$.  In our example, $CONSIST = \frac{1}{1000} (\frac{750}{1} + \frac{200}{2} + \frac{50}{3})$.  In this way, models are rewarded for producing a small number of groups with large sizes, suggesting higher consistency. 

\subsection{Pair-Wise BLEU}\label{sec:alt}

The CONSIST metric we propose has several strengths.  It ensures very fine-grained similarity, which would benefit NLP systems in which MT is part of a larger pipeline.  In such a case, tiny variations could cause large down-stream errors.   CONSIST is also quick to compute since it does not require comparing each hypothesis with each other one separately.  However, we concede that CONSIST may not be best in all settings, especially for domains with longer sentences.

To address this weakness, we also calculate a pairwise-BLEU-based score in which we compute average sentence BLEU between all pairs of hypotheses for a prompt and average across all prompts.  Specifically, for a pair $(X_i,\hat{Y}^{\theta}_i)$:

$$ PWB^{\theta}_i \triangleq \frac{1}{n_i}\sum_{j=1}^{|\hat{Y}^{\theta}_i|}\sum_{k=j+1}^{|\hat{Y}^{\theta}_i|}sBLEU(\hat{y}_i^j,\hat{y}_i^k), $$

$$ n_i = \frac{(|X_i|)(|X_i| - 1)}{2} $$

Thus a model's final PWB score is:

$$ PWB^{\theta} \triangleq \frac{1}{|C|} \sum_{i=1}^{|C|} PWB^{\theta}_i.$$

This metric is like Minimum Bayes Risk \cite{freitag-etal-2022-high} and like that metric, takes significantly longer than CONSIST to compute. It mitigates some of the negatives of CONSIST since it works for long sentences and allows for less-strict similarity evaluation.  

We observe the same consistency patterns from our models using this pairwise score as the patterns drawn from CONSIST shown in Table \ref{tab:pwb}.

\begin{table}[]
    \centering
    \begin{tabular}{c||c|c}
         & CONSIST & PWB \\
         \hline
       m2m  & 54.7 & 58.3 \\
     + 1 most & 47.3 & 53 \\
     + 10 random & 72.3 & 73.2 \\
     + all & 77.6 & 78.3
    \end{tabular}
    \caption{CONSIST scores match patterns with the more robust yet more expensive PWB metric for consistency in pt-en.}
    \label{tab:pwb}
\end{table}

The similarity in consistency scores between these two metrics motivates our usage for the quicker of the two, CONSIST.

\subsection{Synthetic Perturbation}\label{sec:rules}
\subsubsection{Latin Script Perturbations (hu, pt)}
For the Latin-script perturbations, each word (determined by white space) has a uniform random-chance of one of the following perturbations:
\begin{itemize}
    \item No change
    \item Casing e.g. `apple' $\rightarrow$ `Apple'
    \item Insertion e.g. `apple' $\rightarrow$ `appqle'
    \item Deletion e.g. `apple' $\rightarrow$ `aple'
    \item Subsitution e.g. `apple' $\rightarrow$ `apqle'
\end{itemize}

\subsubsection{Japanese Script NAV Perturbation}
For the Japanese-script perturbations, we use rules (which also have uniform chance of being implemented after certain conditions are met).  These rules attempt to create true NAV perturbations, compared to the more-contrived Latin-script perturbations.  Rules include:
\begin{itemize}
    \item Sentence-Ending Emphasis Particles e.g. `' $\rightarrow$ `\begin{CJK}{UTF8}{min}よ\end{CJK}'
    \item Copula removal e.g. `\begin{CJK}{UTF8}{min}です\end{CJK}' $\rightarrow$ `'
    \item Pronoun substitution e.g. `\begin{CJK}{UTF8}{min}私\end{CJK}' $\rightarrow$ `\begin{CJK}{UTF8}{min}俺\end{CJK}'
\end{itemize}

\subsection{Full Mono-Pair Results}

Table \ref{tab:directional} shows our full results from the mono-pair model experiments.  In addition to CONSIST and BLEU, we also obtain a MATCH score, based on average percentage the output sentence matches the reference and NUM, the average number of different translations generated per set of semantically-equivalent NAV perturbations.

\begin{table}[]
\centering
\begin{tabular}{ll|rr|rr}
\multicolumn{2}{l|}{\textbf{hu}-en}  & \multicolumn{2}{l|}{Quality}                         & \multicolumn{2}{l}{Consistency}                     \\
cond                  & \#  & \multicolumn{1}{l}{BLEU} & \multicolumn{1}{l|}{MAT.} & \multicolumn{1}{l}{CONS.} & \multicolumn{1}{l}{NUM} \\ \hline
base                  & 0   & 28.2                     & 11.1                      & 43.1                      & 21.3                    \\ \hline
\multirow{3}{*}{most}  & 1   & 37.4                     & 22.3                      & 44.8                      & 20.5                    \\
                      & 2   & 42.3                     & 26.7                      & 51.3                      & 16.5                    \\
                      & 10  & 45.4                     & 29.6                      & 58.0                      & 12.4                    \\ \hline
\multirow{3}{*}{rand} & 1   & 46.3                     & 29.8                      & 56.3                      & 12.9                    \\
                      & 2   & 46.9                     & 32.0                      & 58.3                      & 11.9                    \\
                      & 10  & 46.6                     & \textbf{32.3}             & 62.9                      & 10.0                    \\ \hline
\multirow{3}{*}{least}  & 1   & 47.2                     & 30.6                      & 56.4                      & 13.0                    \\
                      & 2   & \textbf{48.1}            & 31.4                      & 58.8                      & 11.3                    \\
                      & 10  & 46.9                     & 31.2                      & 62.8                      & 9.9                     \\ \hline
all                   & all & 43.3                     & 26.4                      & \textbf{64.4}             & \textbf{8.4}    

\vspace{.5cm}
\end{tabular}


\begin{tabular}{ll|rr|rr}
\multicolumn{2}{l|}{\textbf{ja}-en}  & \multicolumn{2}{l|}{Quality}                         & \multicolumn{2}{l}{Consistency}                     \\
cond                  & \#  & \multicolumn{1}{l}{BLEU} & \multicolumn{1}{l|}{MAT.} & \multicolumn{1}{l}{CONS.} & \multicolumn{1}{l}{NUM} \\ \hline
base                  & 0   & 29.0                     & 14.5                      & 39.5                      & 46.8                    \\ \hline
\multirow{3}{*}{most}  & 1   & 45.7                     & 27.3                      & 48.9                      & 29.4                    \\
                      & 2   & 45.6                     & 28.5                      & 49.6                      & 28.1                    \\
                      & 10  & 50.5                     & \textbf{33.6}                      & 60.5                      & 16.2                    \\ \hline
\multirow{3}{*}{rand} & 1   & 47.9                     & 30.8                      & 54.3                      & 21.6                    \\
                      & 2   & \textbf{51.0}            & 33.2            & 57.3                      & 18.1                    \\
                      & 10  & 50.0                     & 33.0             & 60.3                      & 15.2                    \\ \hline
\multirow{3}{*}{least}  & 1   & 49.4                     & 31.7                      & 55.1                      & 21.2                    \\
                      & 2   & 48.8            & 32.2                      & 56.8                      & 18.6                    \\
                      & 10  & 50.1                     & 33.1                      & 61.7                      & 13.9                    \\ \hline
all                   & all & 40.8                     & 22.6                      & \textbf{66.4}             & \textbf{9.1}     
\vspace{.5cm}
\end{tabular}

\begin{tabular}{ll|rr|rr}
\multicolumn{2}{l|}{\textbf{pt}-en}  & \multicolumn{2}{l|}{Quality}                         & \multicolumn{2}{l}{Consistency}                     \\
cond                  & \#  & \multicolumn{1}{l}{BLEU} & \multicolumn{1}{l|}{MAT.} & \multicolumn{1}{l}{CONS.} & \multicolumn{1}{l}{NUM} \\ \hline
base                  & 0   & 38.4                     & 12.0                      & 35.3                      & 34.3                    \\ \hline
\multirow{3}{*}{most}  & 1   & 42.0                     & 14.1                      & 34.2                      & 34.0                    \\
                      & 2   & 43.4                     & 16.2                      & 37.5                      & 29.8                    \\
                      & 10  & 52.8                     & 26.1                      & 53.3                      & 16.9                    \\ \hline
\multirow{3}{*}{rand} & 1   & 52.4                     & 26.3                       & 51.6                      & 18.5                    \\
                      & 2   & 54.8            & 26.3             & 55.8                      & 15.7                    \\
                      & 10  & \textbf{56.9}            & 29.0             & 59.2                      & 13.0                    \\ \hline
\multirow{3}{*}{least}  & 1   & 53.8                     & 27.9                      & 52.8                      & 18.2                    \\
                      & 2   & 55.7            & 29.4                      & 55.3                      & 15.3                    \\
                      & 10  & 56.5                     & \textbf{30.0}                      & 59.3                      & 12.8                    \\ \hline
all                   & all & 53.0                     & 26.9                      & \textbf{63.0}             & \textbf{10.0}          
\end{tabular}
\caption{Results for translation quality and consistency for three language directions under different fine-tuning conditions.  Including additional perturbations increases test consistency consistently.}
\label{tab:directional}
\end{table}

\subsection{Multi-pair Model Results}\label{sec:multi}

\begin{figure}[h]
    \centering
    \includegraphics[scale=.7]{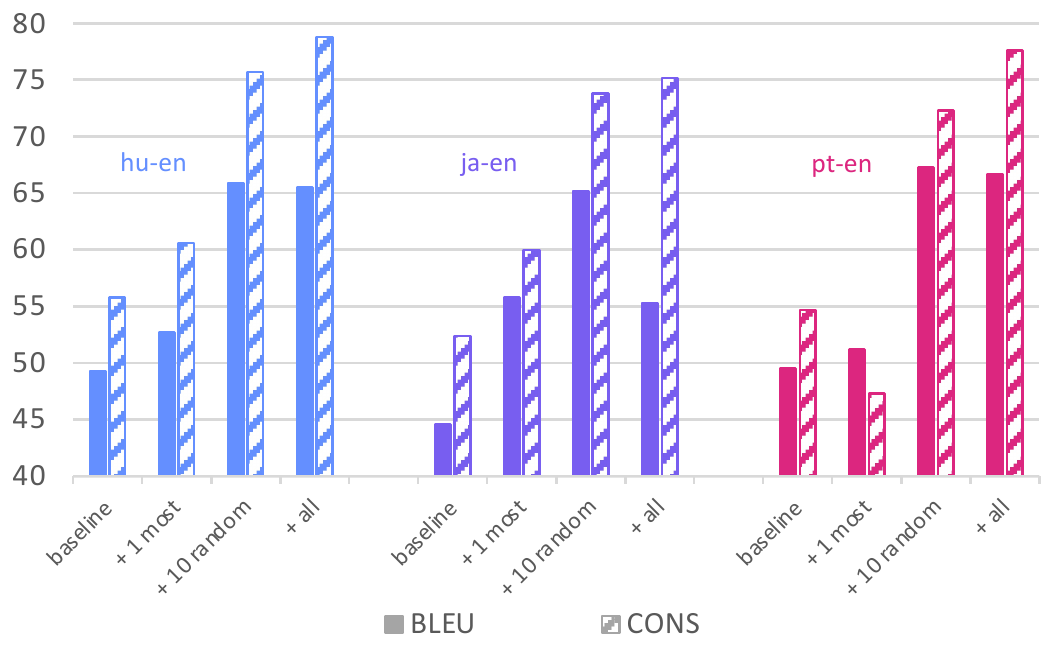}
    \caption{As in the mono-pair translation experiments, we see how `+ 1 most' generally shows improvements over a baseline and `+ 10 random' shows how exposing the model to NAV perturbations further increases both BLEU and CONSIST scores.  While including all NAV perturbations continues to improve CONSIST, we see a trade-off in BLEU.}
    \label{fig:MULT}
\end{figure}

Results from multi-pair experiments in Section \ref{sec:trans} in Figure \ref{fig:MULT}.

\subsection{Quantitative Error}\label{sec:quant}

Getting clear quantitative descriptions for the frequency of specific types of errors seen in baseline models is difficult, but we provide some more information about the analysis.  We find the types of errors mentioned (issues with dropped pronouns, uncommon synonyms, gender ambiguity, etc.) fairly frequently.  Skimming through several prompts, we see these types of erroneous output many times.  

For example, in a prompt with 92 NAV perturbations in Japanese, 52 of the outputs from the baseline model were incorrect.  With a reference “It is a kind of tomato”, erroneous outputs include “Type of tomatoes” “It is like tomatoes” “Something about tomatoes” and “One sort of tomato”; correct outputs include “It is a variety of tomato” “This is a type of tomato” and “It is a tomato species”.  Some of these errors are language specific, since NAV is very dependent on the two languages involved, but as several languages share properties, relevant errors can be seen across languages (e.g. pronoun drop in Japanese, Korean and Portuguese)

\section{Additional Evaluations}

\subsection{Out-of-Domain Performance of NAV-robust Models}\label{sec:ood}

\begin{figure}[h]
    \centering
    \includegraphics[scale=.6]{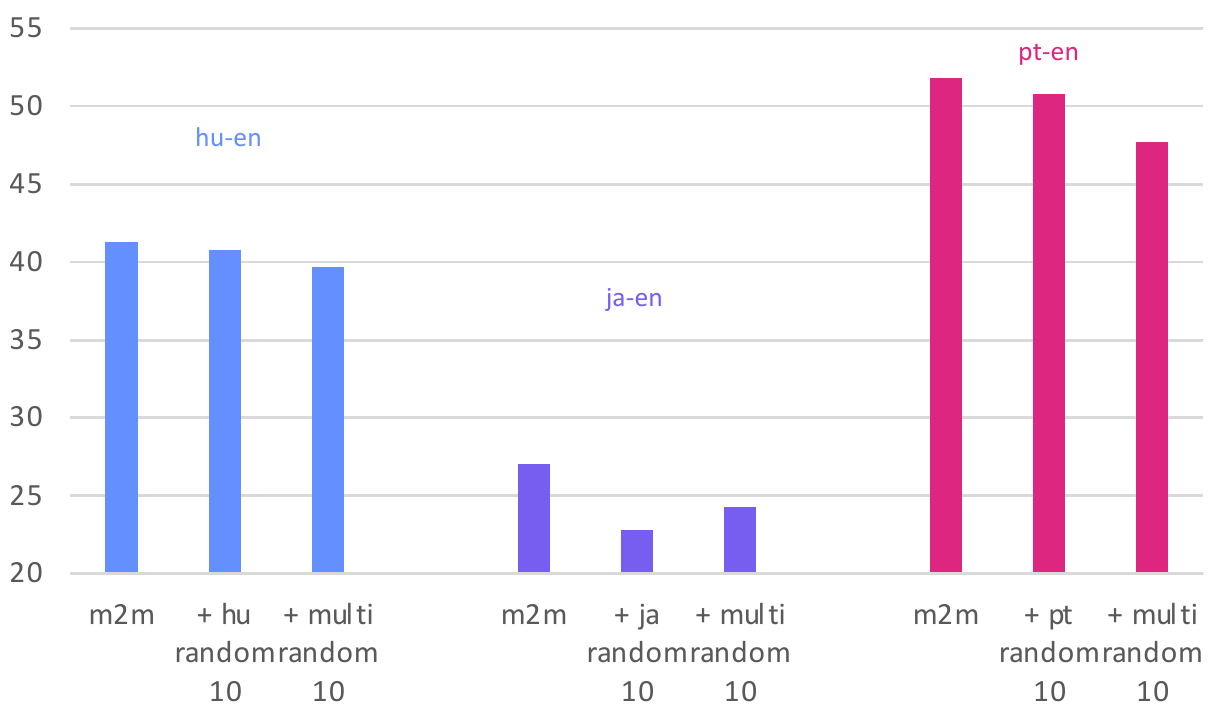}
    \caption{BLEU scores on held-out Tatoeba test sets.  Baseline models perform better than NAV-robust models.}
    \label{fig:tato}
\end{figure}
\begin{figure}[h]
    \centering
    \includegraphics[scale=.45]{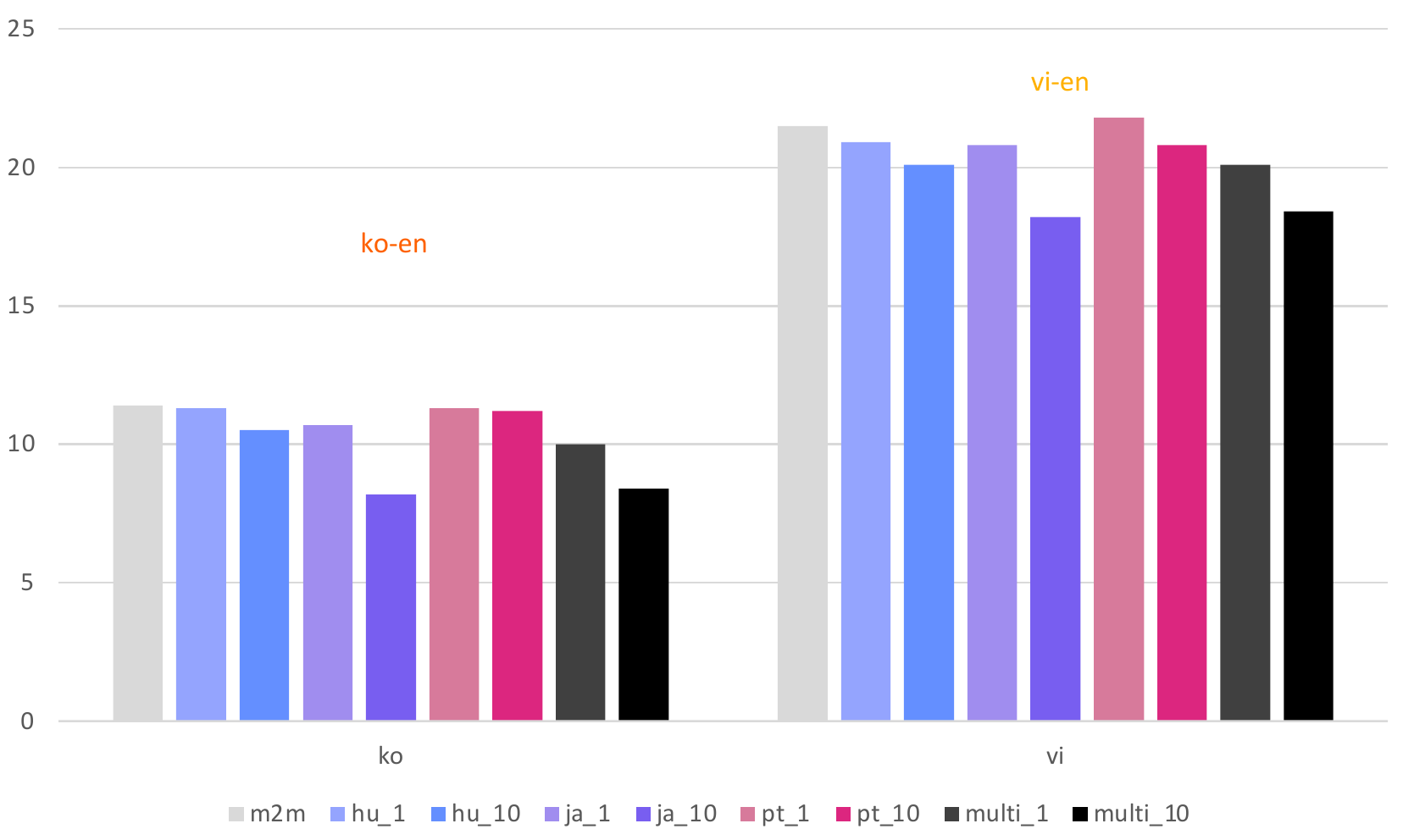}
    \caption{BLEU scores on held-out OPUS-100 test sets.  Baseline models perform better than most NAV-robust models.}
    \label{fig:opus}
\end{figure}

To further investigate behavior of NAV-robust models, we run experiments to see the effect on performance in OOD tasks.  For \{hu,ja,pt\}-en we use a held-out test split of the Tatoeba corpus.  Our results are shown in Figure \ref{fig:tato}.  For \{ko,vi\}-en we use the official OPUS-100 test set with results shown in Figure \ref{fig:opus}.

From these experiments, we see that generally NAV-robustification slightly worsens OOD performance.  While these results appear negative, they are not too surprising considering common trends in fine-tuning and robustification \cite{he-etal-2021-analyzing}.  Often, fine-tuning can cause some forgetting and models with higher robustness perform worse when evaluated on original, un-noisy input. Correcting this widespread behavior is beyond the scope of this work.

While it's clear that improving NAV robustness has shown a decrease in `regular' MT BLEU, there are many differences between our NAV evaluation settings and our experiment settings in this section.  For example, these experiments are not just 1) out of domain, they also 2) no longer have a robustness component to them because we aren't evaluating the models on several perturbations (NAV or otherwise) of input sentences.  The decrease in BLEU may be due to either or both of these properties, which our next subsections attempt to disentangle.

\subsection{NAV-robust Models on Other Types of Noise}\label{sec:rob}
Our next set of experiments in this section looks at measuring NAV-robustness transfer.  By this we seek to answer the question \emph{does a model fine-tuned deliberately for NAV robustness exhibit robustness to non-NAV-related noise?}  For this, we use an existing MT test set designed to include noisy input text to challenge models' robustness, MTNT (Machine Translation of Noisy Text) \cite{michel-neubig-2018-mtnt}.

\begin{figure}[h]
    \centering
    \includegraphics[scale=.6]{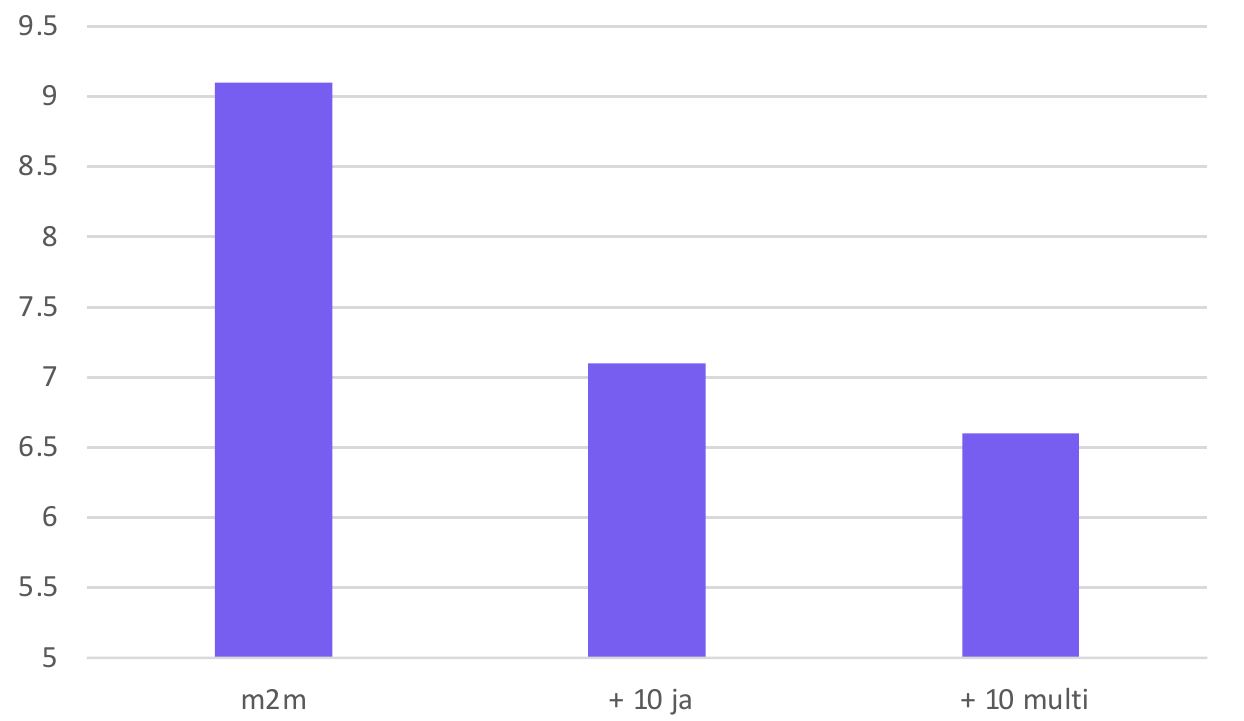}  
    \caption{NAV-robust models perform worse than a baseline model in MTNT.  MTNT is a difficult task (highlighted by the very low BLEU scores) and includes greater and more varied noise than within the NAV boundaries.}
    \label{fig:mtnt}

\end{figure}

We simply evaluate our baseline and select NAV-tuned models on the ja-en test split of MTNT.  Our results are shown in Figure \ref{fig:mtnt}.  We see that our baseline performs better than our NAV-tuned models, suggesting zero-shot transfer of robustness may not be possible in this way.  The NAV perturbations which our models were trained to be robust to do not overlap with many of the types of noise in MTNT, which resembles less-standard `internet-speech'.

Again, we have another problem with this evaluation in that the models are not only tested against new classes of perturbations but also in a new domain.  One way of testing performance on new classes of perturbation but staying in-domain is to use our synthetic augmentation scripts to create a synthetic robustness test set.

\begin{figure}[h]
    \centering
    \includegraphics[scale=.6]{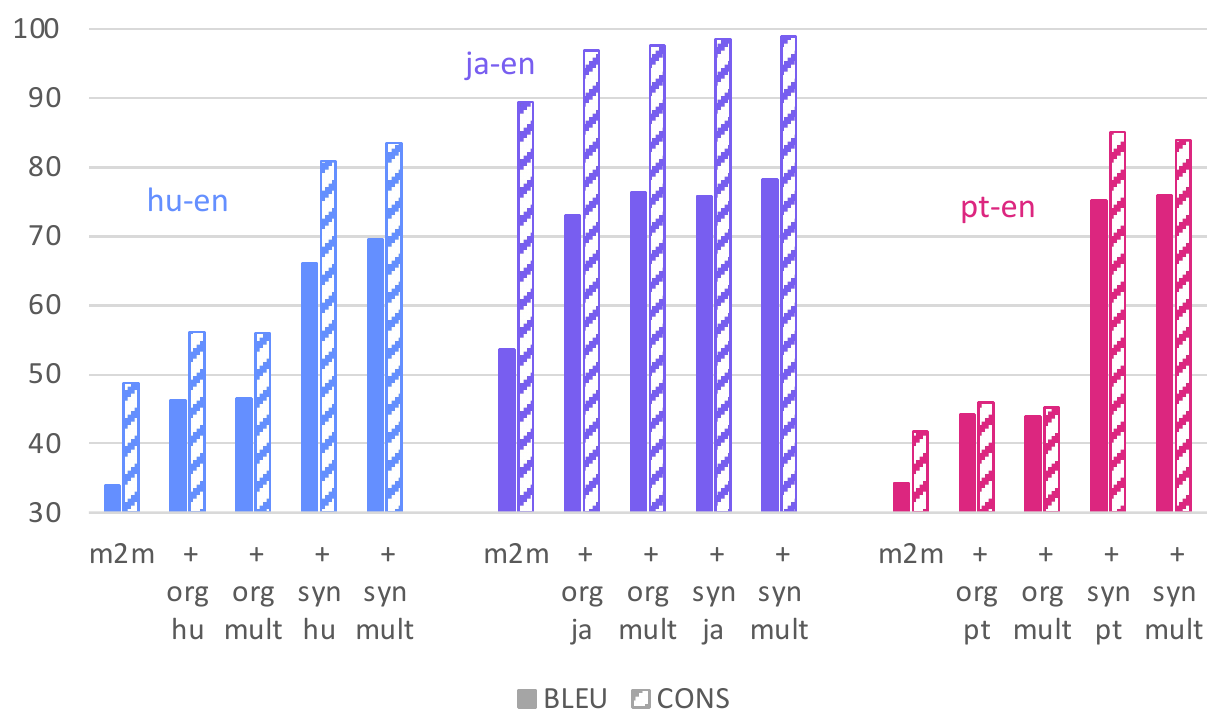}
    \caption{BLEU and CONSIST scores for organic vs. synthetic perturbations during fine-tuning on synthetic STAPLE test sets.  Synthetic fine-tuning unsurprisingly improves synthetic test performance, but we also see how organic NAV fine-tuning performs fairly well on rule-based synthetic perturbation which aims to emulate NAV (ja-en) compared to random character-alteration synthetic test sets (\{hu,pt\}-en).}
    \label{fig:synt}
\end{figure}

We use the same scripts but apply them to the test splits of \{hu,ja,pt\} STAPLE.  In this way, we can probe transferability of NAV robustness by evaluating our NAV-robust models on these synthetic test sets.  We also evaluate the models fine-tuned on synthetic data for comparison.  Results are shown in Figure \ref{fig:synt}.  

In general, the synthetic-tuned models perform better on these synthetic test evaluations.  However, the organic NAV-tuned models do exhibit some improved robustness compared to a baseline model.  The NAV perturbations these models are fine-tuned on rarely overlap with the types of synthetic perturbations performed on the new \{hu,pt\}-en sets, explaining the only slight transferability.  Organic NAV perturbations help more on our synthetic ja-en test set, likely because the ja-en test set attempts to synthetically create NAV perturbations, thus imitating the organic fine-tuning data more closely.

\subsection{In-Domain MT}\label{sec:in}
One possible source of confusion in our NAV robustness evaluation could be the fact that it is abnormal to have hundreds of test examples with the same reference translation.  Perhaps BLEU is not as reliable in such conditions.  To account for this, we calculate BLEU on in-domain data by using our `1-most' and `1-least' STAPLE test sets.  This returns the task to single-pair examples while focusing on common and uncommon examples, respectively.

\begin{figure}[h]
    \centering
    \includegraphics[scale=.6]{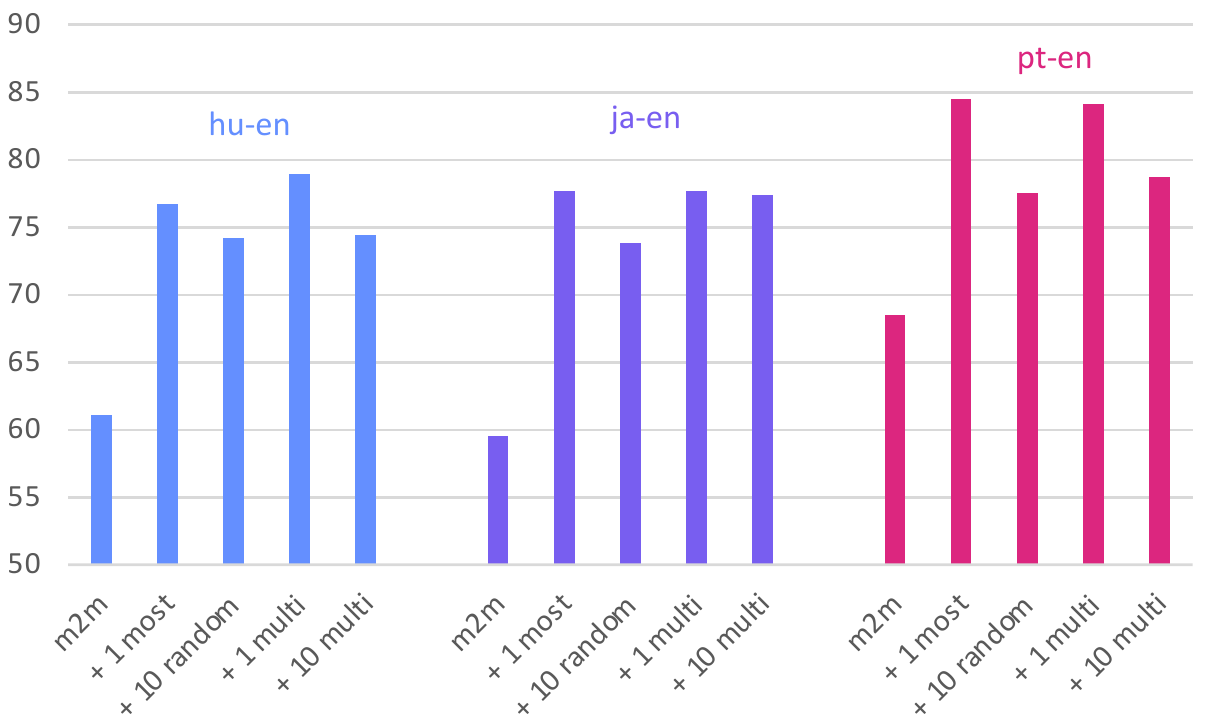}
    \caption{BLEU scores evaluating on 1-most STAPLE test sets.  1-most fine-tuning improves performance the most.}
    \label{fig:most}
\end{figure}

\begin{figure}[h]
    \centering
    \includegraphics[scale=.6]{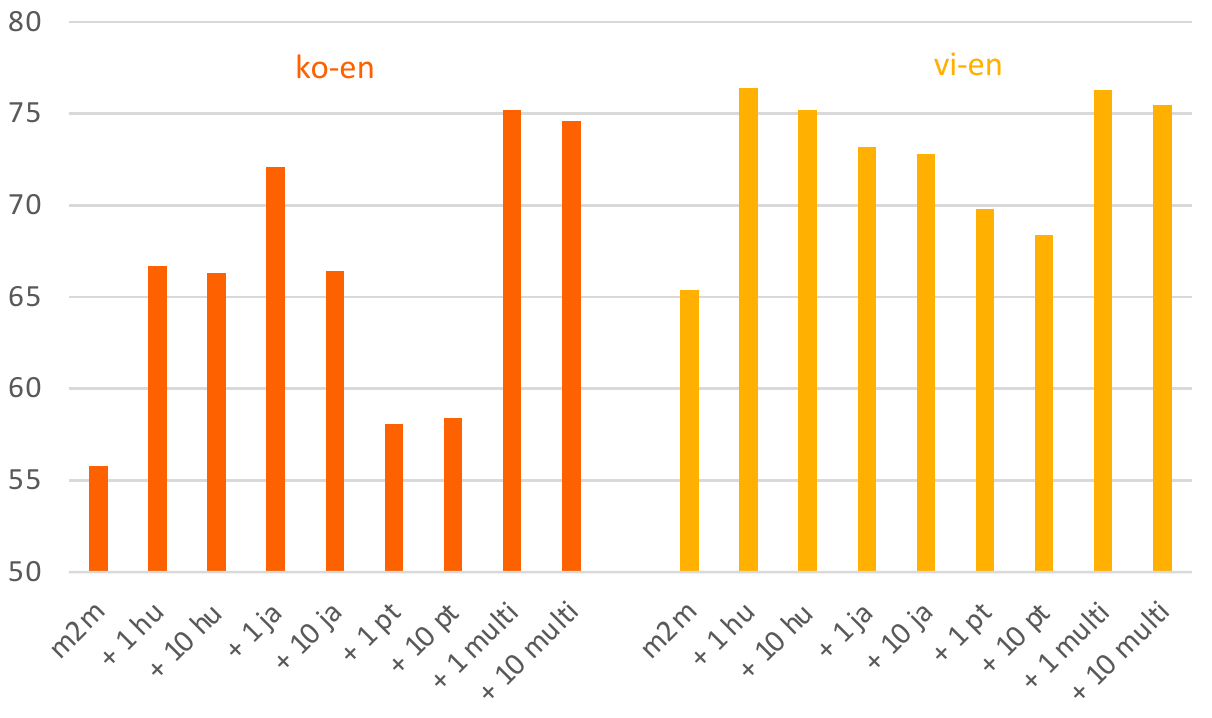}
    \caption{BLEU scores evaluating on 1-most STAPLE test sets for 0-shot language transfer.  1-most fine-tuning improves performance over baseline and 10-random models.  We see similar language differences with ja-en improving ko-en more compared to hu-en with vi-en.}
    \label{fig:most2}
\end{figure}

The results for `1-most' are shown in Figures \ref{fig:most} and \ref{fig:most2}. In our NAV experiments from Section 4, the ``10 random'' models outperform the ``1 most'' models, but here we see the opposite.  This is not surprising as the `1-most' fine-tuning is most similar to this `1-most' evaluation.  Also, this evaluation no longer contains a robustness aspect as each example pair uses a commonly-occurring source sentence.

\begin{figure}[h]
    \centering
    \includegraphics[scale=.6]{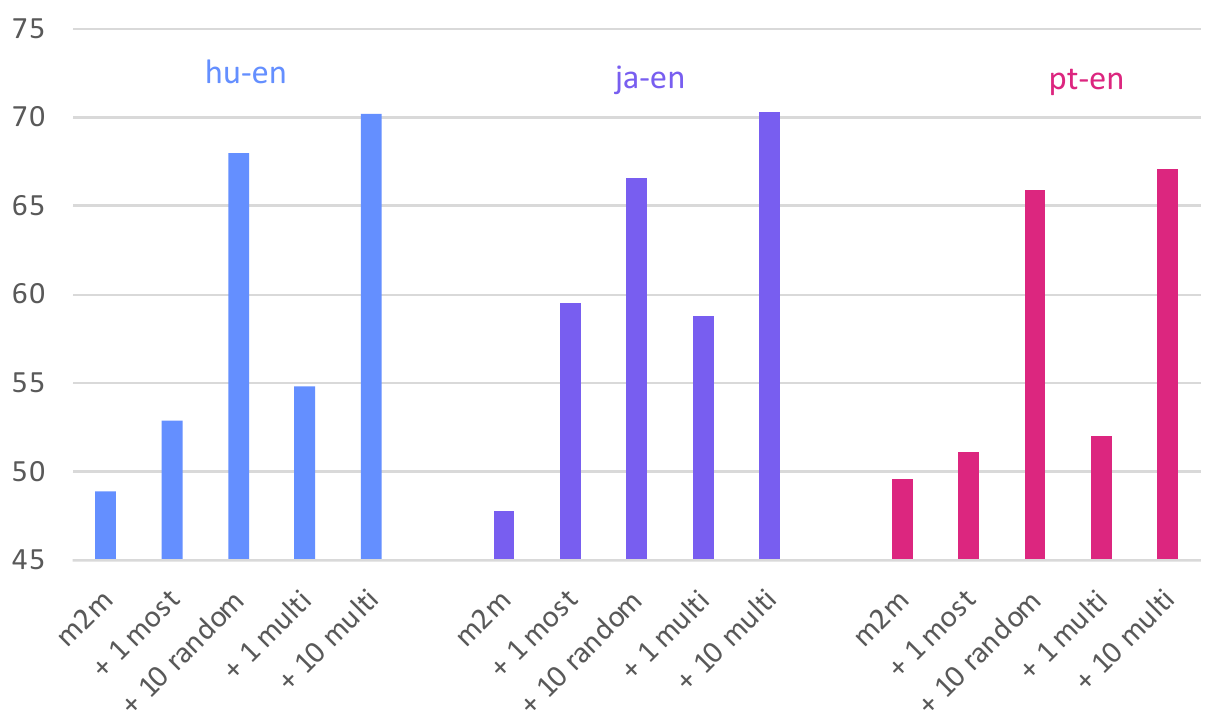}
    \caption{BLEU scores evaluating on 1-least STAPLE test sets.  10-random fine-tuning improves performance the most.}
    \label{fig:least}
\end{figure}

\begin{figure}[h]
    \centering
    \includegraphics[scale=.6]{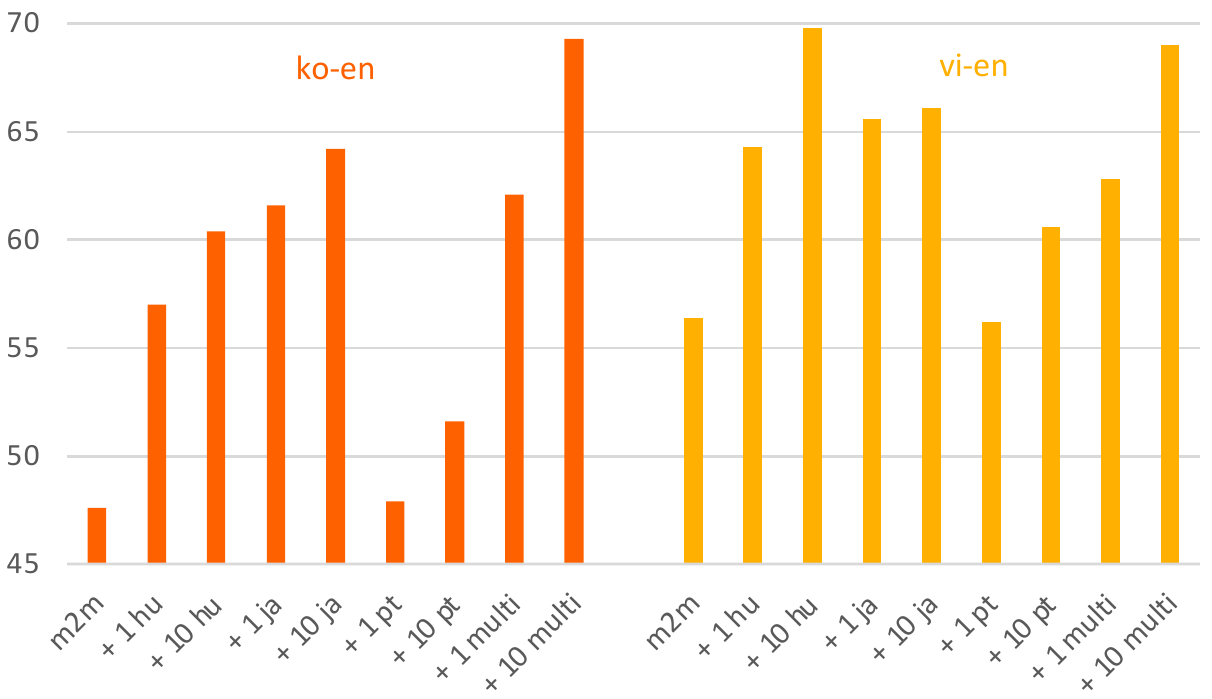}
    \caption{BLEU scores evaluating on 1-least STAPLE test sets for 0-shot language transfer.  10-random fine-tuning improves performance over baseline and 1-most models.  We see language differences with ja-en improving ko-en more compared to hu-en with vi-en.}
    \label{fig:least2}
\end{figure}

This raises the concern that it could be, in fact, that 10-random has higher BLEU on our robustness experiments because of the property of having hundreds of variations for each example.  To disentangle this possible explanation, we also evaluate our models on a ``1-least'' STAPLE set. We source the \emph{least} common source sentence for each example.  This helps ensure the test inputs will exhibit more difficult NAV perturbations, thus serving as a better NAV-robustness test set without the added property of having several variations from each translation pair.  The results are shown in Figures \ref{fig:least} and \ref{fig:least2}.

Our results now reflect those from our robustness experiments with ``10-random'' fine-tuning consistently performing better than ``1-most''.  This suggests that the models showing highest NAV robustness by our evaluations also show higher robustness in a more standard MT evaluation on NAV-noisy data.

\end{document}